\documentclass[conference]{IEEEtran}
\IEEEoverridecommandlockouts
\usepackage{cite}
\usepackage{amsmath,amssymb,amsfonts}
\usepackage{algorithmic}
\usepackage{graphicx}
\usepackage{textcomp}
\usepackage{float}
\usepackage{xcolor}
\usepackage{url}
\usepackage{tikz}
\usepackage{pgfplots}
\pgfplotsset{compat=1.17}
\usepackage{subcaption}

\begin{document}

\title{Arithmetic-Intensity-Aware Quantization}

\author{\IEEEauthorblockN{Taig Singh}
\IEEEauthorblockA{\textit{Harvard College}\\
Cambridge, MA, USA \\
taig\_singh@college.harvard.edu}
\and
\IEEEauthorblockN{Shreshth Rajan}
\IEEEauthorblockA{\textit{Harvard College}\\
Cambridge, MA, USA \\
shreshthrajan@college.harvard.edu}
\and
\IEEEauthorblockN{Nikhil Jain}
\IEEEauthorblockA{\textit{Harvard College}\\
Cambridge, MA, USA \\
nikhiljain@college.harvard.edu}
}

\maketitle

\begin{abstract}

As modern neural networks become increasingly memory-bound, inference throughput is limited by DRAM bandwidth rather than compute. We present Arithmetic-Intensity-Aware Quantization (AIQ), a mixed precision quantization framework that chooses per-layer bit-widths to maximize arithmetic intensity (AI) while minimizing accuracy loss. AIQ is a post-training quantization method that uses search algorithms over per-layer quantization schemes to minimize a weighted loss over AI and accuracy. On ResNet-20/CIFAR-10, AIQ increases AI by about 50\% over an FP32 baseline while keeping test accuracy within $~1$ percentage point, and outperforming global uniform quantization schemes. On a memory-bound MobileNetV2 architecture, AIQ configurations give a 1.66x higher throughput than the FP32 baseline while keeping test accuracy within $1$ percentage point. We also find that AIQ naturally quantizes larger layers more aggressively.

\end{abstract}

\begin{figure}[H]
    \centering
    \begin{tikzpicture}[
        font=\scriptsize,
        box/.style={
            rectangle,
            draw,
            rounded corners,
            align=center,
            minimum width=3.4cm,
            minimum height=0.8cm
        },
        resbox/.style={
            rectangle,
            draw,
            rounded corners,
            align=left,
            minimum width=3.7cm,
            minimum height=1.1cm
        }
    ]

    % Column titles
    \node[font=\scriptsize\bfseries] (aiqtitle) at (-2.2,1.6) {AIQ};
    \node[font=\scriptsize\bfseries] (restitle) at ( 2.2,1.6) {Results};

    % ---------- Left column: AIQ flowchart ----------
    \node[box] (fp32)   at (-2.2, 1.0) {Trained FP32 model};
    \node[box] (search) at (-2.2, -0.1) {Search over per-layer\\bit-widths\\(Greedy \& Coord. Descent)};
    \node[box] (loss)   at (-2.2,-1.3) {Minimize AIQ loss\\$\mathcal{L}(q;\lambda)$ via\\global AI \& accuracy};
    \node[box] (scheme) at (-2.2,-2.4) {AI-aware\\quantization scheme $q$};

    \draw[->] (fp32) -- (search);
    \draw[->] (search) -- (loss);
    \draw[->] (loss) -- (scheme);

    % ---------- Right column: results summary ----------
    \node[resbox] (resnet) at (2.2, 0.4)
        {\textbf{ResNet-20 / CIFAR-10 (T4 GPU)}\\
         -- $\sim$51--52\% higher AI vs FP32\\
         -- $\leq$1\% absolute accuracy drop\\
         -- Beats uniform INT8/INT4 (AI$\times$Acc)};

    \node[resbox] (mobilenet) at (2.2,-1.2)
        {\textbf{MobileNetV2 / CIFAR-10 (Xeon, 2 vCPUs)}\\
         -- Memory-bound model \\
         -- 1.66$\times$ throughput vs FP32\\
         -- Higher throughput than random\\
         \hspace*{0.7em}INT4 on 10 layers};

    \end{tikzpicture}
    \caption{Overview of Arithmetic-Intensity-Aware Quantization (AIQ). 
    Left: starting from a trained FP32 model, search over per-layer bit-widths 
    and minimizes a joint AI/accuracy loss to produce an AI-aware quantization scheme $q$. 
    Right: on ResNet-20/CIFAR-10 and MobileNetV2/CIFAR-10, 
    AIQ significantly increases AI and throughput with 
    minimal accuracy loss compared to FP32 and uniform baselines.}
    \label{fig:intro-aiq-overview}
\end{figure}
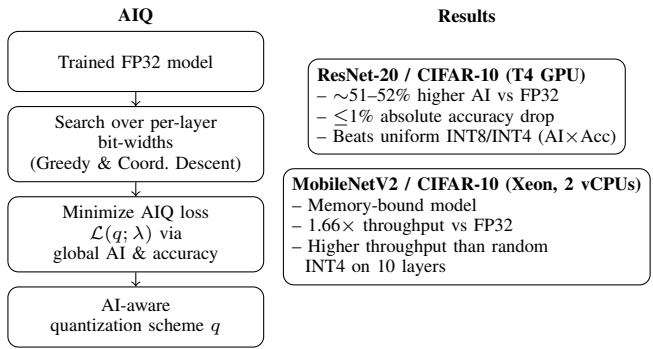

Deep neural networks are now used in many applications. As these workloads have grown in size and scale, inference efficiency has become more and more important to optimize. Many contemporary models have become memory-bound rather than compute-bound, and the Roofline model formalizes this bottleneck by relating the achievable performance of a model to arithmetic intensity (AI). For a given machine, kernels below a certain AI are memory-bound, and for such kernels increasing AI directly translates to a higher throughput \cite{b1}. Quantization is one of the most common approaches for improving inference efficiency. Prior work has developed post-training quantization and quantization-aware training schemes that preserve accuracy at 8-bit and lower precisions \cite{b2}.

However, most quantization methods only optimize model size or directly optimize latency on a target device. Additionally, quantization is typically applied uniformly across models and do not emphasize per-layer quantization choices. Thus, such schemes can be suboptimal on memory-bound workloads as they can over-quantize layers whose memory traffic is not a bottleneck and under-quantize memory-bound layers. At the same time, it is known that different layers have varying sensitivity to quantization. 

This work introduces Arithmetic-Intensity-Aware Quantization (AIQ), a framework that views per-layer quantization as a resource-allocation problem in the Roofline model's space. Starting from a trained full-precision model, we assign each layer a quantization level, forming a scheme over all layers. For any scheme we compute (i) the model's global arithmetic intensity analytically from its FLOPs and memory traffic, and (ii) measure the mode's accuracy empirically. We then define an AI-aware objective that takes into account both the scheme's resulting AI and accuracy.

Our approach and results are summarized in Figure 1. We find that relative to the FP32 baseline and uniform INT8/INT4 baselines, AIQ achieves similar AI gains with significantly smaller accuracy loss, and this is reflected in significant throughput increases.

Section 2 describes our overall approach, including a description of AIQ objective and search algorithms. Section 3 describes our hardware and the models that we use, including ResNet-20 and MobileNet trained on CIFAR-10. Section 4 enumerates our results, showing an empirical validation of AIQ as we achieve large AI gains with a minimal accuracy dropoff compared to baselines. Section 5 covers related works and Section 6 contains the conclusion of our research.

\section{Approach}

AIQ is novel in making model-level AI a first-class optimization objective, and our layerwise AI-accuracy profiling directly measures how quantizing each layer in isolation moves the model on the accuracy-AI plane. Unlike prior work, we validate that AI gains translate into real throughput improvements on memory-bound models. For details regarding prior work, please see Related Work.

\subsection{Problem Formulation}

Starting from a trained full-precision model, AIQ assigns an integer bit-width $q_i \in Q = \{FP32, INT8, INT4\}$ to each layer $i$, forming a quantization scheme $q = (q_1, \dots, q_L)$ across the layers of the model. For any configuration $q$, we compute the global arithmetic intensity $AI(q)$ and model test accuracy loss $AccLoss(q)$. The AIQ loss function is then
$
\mathcal{L}(q;\lambda) = -\lambda AI (q) + (1-\lambda)AccLoss(q)
$
where $\lambda \in [0, 1]$ trades off AI versus accuracy. Note that by sweeping $\lambda$ we can generate a family of quantization scheme that trace an accuracy-AI Pareto frontier. Minimizing $\mathcal{L}$ is equivalent to maximizing AI and maxmizing accuracy with a relative weight determined by $\lambda$. The secret weapon in this approach is optimizing the layerwise AI-accuracy tradeoff. Intellectually, this connects quantization with Roofline analysis: quantization reduces memory traffic and thus moves layers from memory-bound toward compute-bound regions. The empirically measured accuracy captures how much such moves cost in task performance.

\subsection{Search Over Mixed-Precision Schemes}

Directly optimizing $\mathcal{L}(q;\lambda)$ over all bit-width combinations is combinatorial. We present two efficient search algorithms below that operate on the full configuration $q$ that we have used in implementation.

\textbf{Greedy search}: We initialize with the all-FP32 configuration $q^0 = q^{FP}$. At iteration $t$, we consider local moves of the form ``reduce bit-width of unquantized layer $i$ from its current value to a lower value $b \in Q$'' to obtain candidate configurations $q^{t, i, b}$. We then compute the change 
$
\Delta \mathcal{L}_{i, b} = \mathcal{L}(q^{t, i, b}; \lambda) - \mathcal{L}(q^{t}; \lambda)
$
for each choice of $i, b$. We pick the move with the largest improvement $i^*, b^* = \arg \min_{i, b} \Delta \mathcal{L}_{i, b}$ and if $\Delta \mathcal{L}_{i^*, b^*} < 0$ we update $q^{t+1} = q^{t, i, b}$. We execute the same number of iterations as there are layers.

\textbf{Coordinate descent}: We initialize with the all-FP32 configuration $q^0 = q^{FP}$. We iterate over all the layers $i = 1, \dots, L$, and for each layer, we hold the other bit-widths fixed and search over $Q$ for the current layer:
$
q_i^{t+1} = \arg \min_{b \in Q} \mathcal{L}(q^t_{-i}, b; \lambda)
$
where $q^t_{-i}$ denotes the current bits for all other layers. After one pass over all layers, we repeat until there is not an updates that reduces $\mathcal{L}$.

In practice, note that both of the above algorithms are cheap because AI can be recomputed analytically, and evaluation runs use small validation subsets during search.

\section{Implementation}

We conduct four experiments using CIFAR-10 models implemented in PyTorch. For all experiments, we hold $\lambda$ fixed at $0.9$. We enumerate the experiments, the models used, and the hardware as follows: (1) Our initial case study uses a standard ResNet-20 architecture trained on CIFAR-10, trained to $\sim$$92\%$ test accuracy. We quantize a single layer at a time and calculate AI and test accuracy each time, in order to justify that AI and accuracy are impacted differently by layer-quantization, so quantization across all model weights is suboptimal. All runs use an NVIDIA T4 GPU. (2) On the same ResNet-20/T4 setup, we intantiate AIQ with allowed bit-widths FP32, INT8, and INT4. Using the Greedy Search and Coordinate Descent algorithms, we generate empirical results for AIQ on accuracy and AI as compared to baselines of uniform model weight quantization. (3) To test whether AI gains translate into real performance improvements, we fine-tune a MobileNetV2 on CIFAR-10 and deploy it on an Intel Xeon CPU instance with 2 virtual CPUs, chosen to make inference memory‑bound. We run the same AIQ search procedure, and measure end-to-end throughput using wall-clock timing, and we compare this to baselines. (4) Finally, we probe structural heuristics for AIQ that heavier layers and later layers tend to be quantized more aggressively. We do so by running AIQ on variants of the ResNet-20 model, such as a heavy-to-light model and a uniform-size model.

\section{Results}

\subsection{A Case Study with ResNet-20 on CIFAR-10}

For the aforementioned case study, where we quantize only a single layer at a time, please find the corresponding plot of AI and accuracy as the quantized (to INT4) layer changes in Figure 2. Note that AI is impacted differently by layer-quantization, and so we conclude that quantization across all model weights is suboptimal. Quantizing a single layer has a clear theoretical effect on AI but an unpredictable empirical effect on accuracy, motivating per-layer analysis.

% Figure 1: Layerwise AI/accuracy (INT4 per-layer case study)
\begin{figure}[!htbp]
    \centering
    \includegraphics[width=0.8\linewidth]{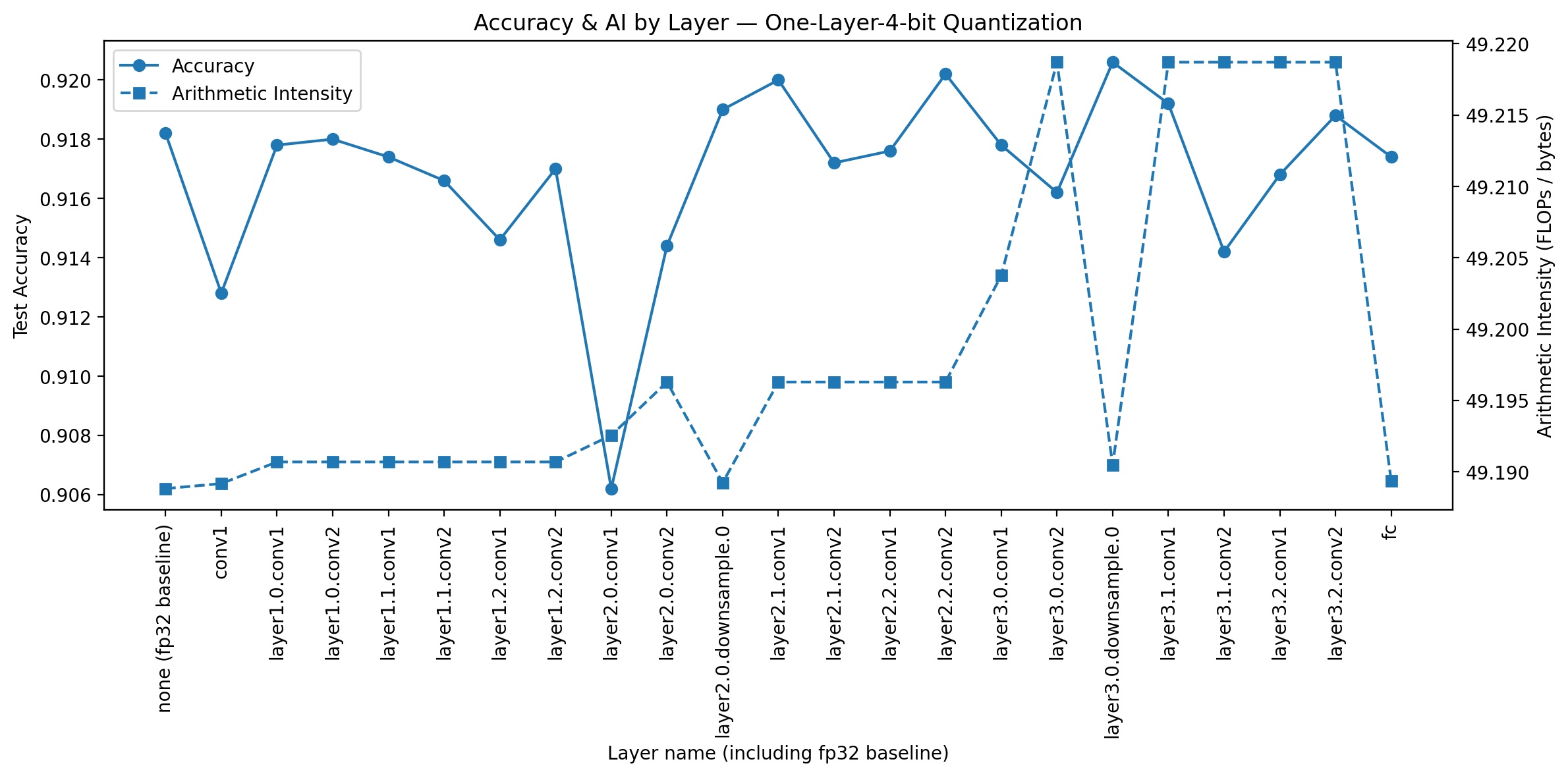}
    \caption{Arithmetic intensity and test accuracy when quantizing a single ResNet-20 layer to INT4, varying the quantized layer.}
    \label{fig:ai-acc-layerwise}
\end{figure}

\subsection{Analysis of Greedy Algorithm}

Figure 3(a) shows a combined performance score defined as Arithmetic Intensity multiplied by test accuracy. AI-Aware quantization achieves the highest score, outperforming both uniform INT8 and INT4. Figure 3(b) shows the accuracy–AI Pareto frontier for the same configurations. FP32 has low AI but high accuracy. INT8 and INT4 move right along the AI axis, with INT4 losing more accuracy. AI-Aware lies on the Pareto frontier, achieving the highest AI while maintaining accuracy above 91 percent. This demonstrates that the greedy algorithm successfully identifies a mixed-precision configuration that improves efficiency without sacrificing too much accuracy.

\begin{figure}[!htbp]
    \centering
    \begin{subfigure}[b]{0.48\linewidth}
        \centering
        \includegraphics[width=\linewidth]{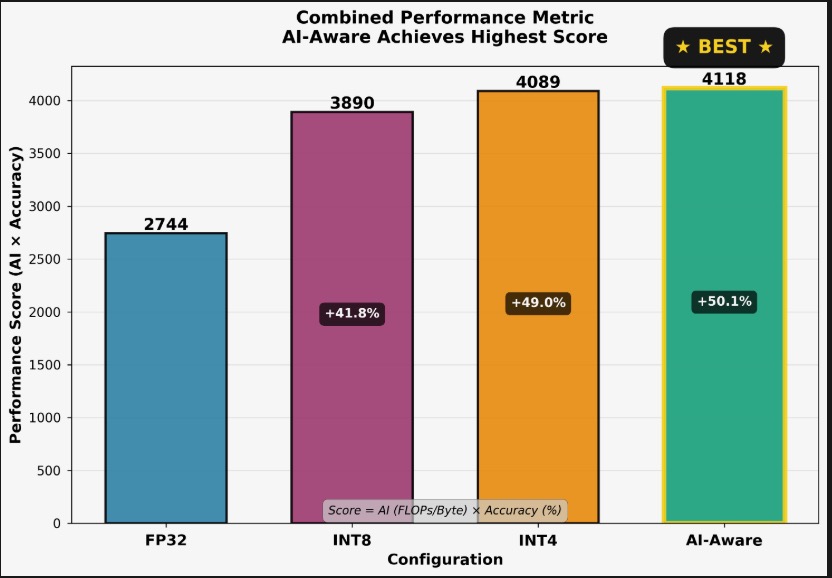}
        \caption{AI (FLOPs/Byte) $\times$ accuracy (\%) for quantization schemes on ResNet-20. AIQ in green.}
        \label{fig:greedy-combined-score}
    \end{subfigure}
    \hfill
    \begin{subfigure}[b]{0.48\linewidth}
        \centering
        \includegraphics[width=\linewidth]{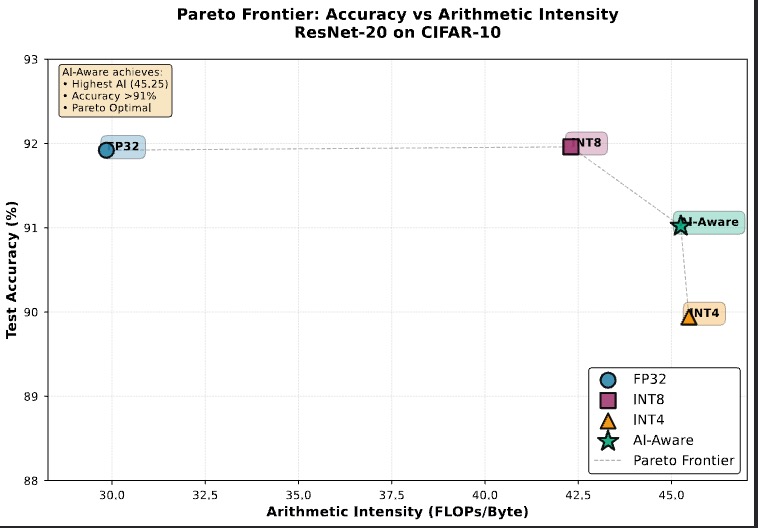}
        \caption{Accuracy--arithmetic intensity Pareto frontier for ResNet-20 under different quantization schemes, including AIQ in green.}
        \label{fig:greedy-pareto}
    \end{subfigure}
    \caption{Greedy AIQ results on ResNet-20: combined performance score and accuracy--AI Pareto frontier.}
    \label{fig:greedy-overall}
\end{figure}

\subsection{Analysis of Coordinate Descent Algorithm}

Figure 4 shows how each quantization method trades off accuracy and arithmetic intensity. Uniform INT8 improves AI by about 42 percent with virtually no accuracy loss, while uniform INT4 achieves the highest AI among the uniform schemes but drops accuracy by nearly 2 percent. The coordinate descent method produces a mixed-precision configuration with AI comparable to INT4 but with much smaller accuracy loss, staying within 1 percent of the FP32 baseline. Its results closely match those of the greedy algorithm, indicating that both search methods converge to similar high-AI, low-loss solutions. 

% Figure 4: Coordinate descent comparison table as image
\begin{figure}[!htbp]
    \centering
    \includegraphics[width=0.8\linewidth]{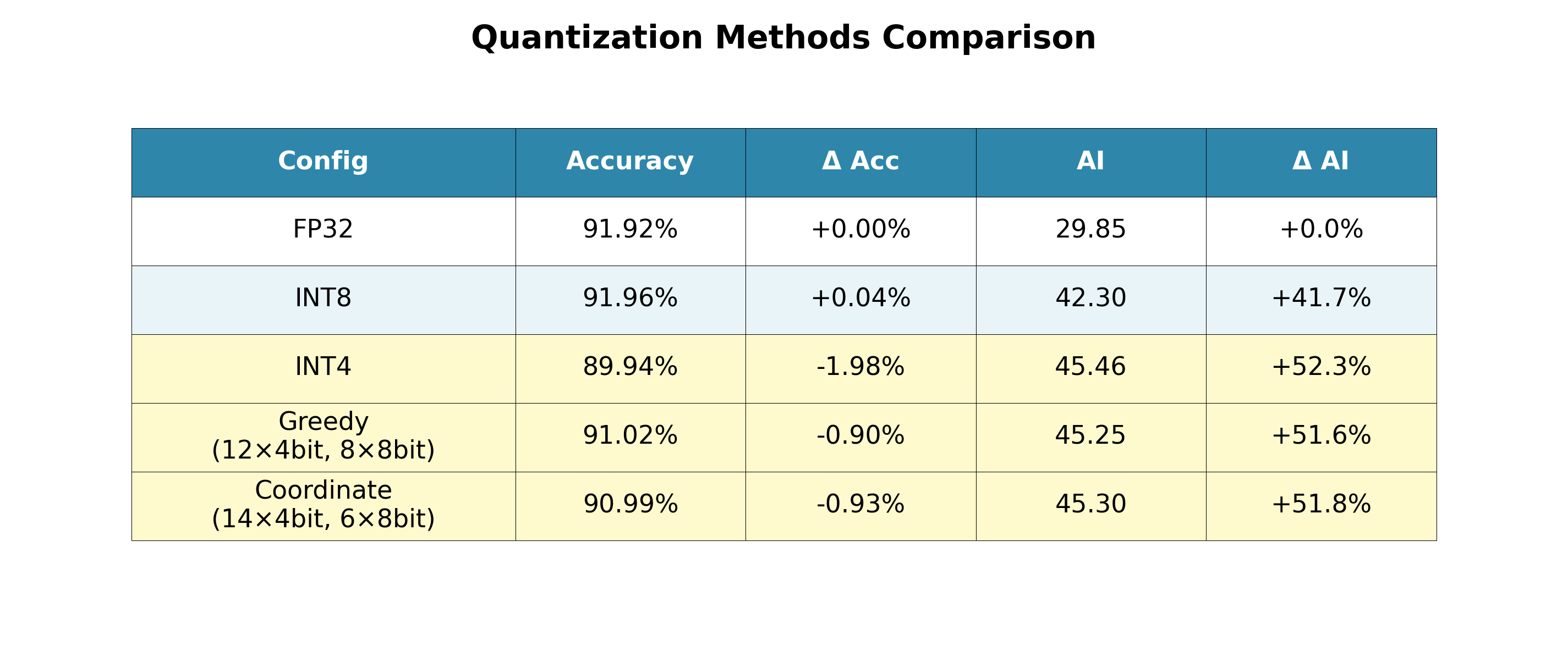}
    \caption{Accuracy and AI comparison across quantization schemes, including coordinate descent AIQ configuration.}
    \label{fig:coord-descent-table}
\end{figure}

\subsection{Analysis of AIQ Throughput Effects}

For the MobileNetV2 throughput study on the Xeon CPU, AIQ delivers clear performance gains while keeping the accuracy loss minimal. The FP32 baseline gives an 82.05\% accuracy at 1396.5 images/s. We run the Greedy Search algorithm for quantizing layers to INT8, and AIQ quantizes 12 layers to INT8, achieving a slightly reduced accuracy of 81.11\% but with a significantly increased throughput of 2320.6 images/s. Since AIQ chose 12 layers to quantize, we also created another baseline to compare AIQ to by randomly quantizing 12 layers to INT8, and this achieves an accuracy of 81.72\% at a lower throughput of 2170.1 images/s. Thus, with minimal accuracy dropoff, AIQ achieves a $66\%$ increase in throughput over the original model.

\subsection{Analysis of Initial Hypotheses}

After our initial experiments, we hypothesized that layers with more channels and those later in the network would be prime candidates for aggressive quantization. To disentangle depth from width, we built HeavyEarlyResNet20, reverses the channel progression of ResNet20 by assigning channel groups of $(64, 32, 16, 16)$ instead of $(16, 16, 32, 64)$, and PlainConvNet20, where all 20 layers have 64 channels, motivated by our observation that later 32/64-channel layers in ResNet-20 were quantized most often. HeavyEarlyResNet20 shows that wide layers are consistently quantized first regardless of depth or accuracy sensitivity, indicating that channel count is the dominant driver. PlainConvNet20 reveals that accuracy sensitivity still matters as layers that are less sensitive are quantized more but channel count often overrides sensitivity, as many wide yet sensitive layers are still quantized. Please find the corresponding per-layer sensitivity and bitwidths in Table 1.

\begin{table}[!htbp]
\tiny
\setlength{\tabcolsep}{1.5pt}
\centering
\caption{Per-layer sensitivity and final bitwidths for models.}
\begin{minipage}{0.49\columnwidth}
\centering
\textbf{PlainConvNet20}\\[2pt]
\begin{tabular}{lccc@{\hspace{3pt}}lccc}
\hline
Lyr & 8b & 4b & BW & Lyr & 8b & 4b & BW\\
\hline
cls & -0.02 &  0.00 & 32 & f30 & -0.01 & -0.01 & 32\\
f0  &  0.02 &  0.03 & 32 & f33 &  0.00 &  0.03 &  4\\
f3  & -0.04 &  0.20 & 32 & f36 & -0.01 & -0.16 &  4\\
f6  &  0.00 & -0.07 &  4 & f39 & -0.01 &  0.09 &  4\\
f9  &  0.00 & -0.04 &  4 & f42 & -0.01 & -0.05 & 32\\
f12 &  0.04 & -0.07 &  4 & f45 &  0.01 &  0.06 & 32\\
f15 &  0.04 & -0.03 &  4 & f48 &  0.00 & -0.08 & 32\\
f18 & -0.01 &  0.00 &  4 & f51 & -0.01 & -0.02 & 32\\
f21 &  0.00 & -0.04 & 32 & f54 &  0.00 & -0.06 &  4\\
f24 &  0.02 &  0.15 &  4 & f57 &  0.01 &  0.01 &  4\\
f27 &  0.02 & -0.02 &  4 &     &       &       &   \\
\hline
\end{tabular}
\end{minipage}%
\hfill
\begin{minipage}{0.49\columnwidth}
\centering
\textbf{HeavyEarlyResNet20}\\[2pt]
\begin{tabular}{lccc@{\hspace{3pt}}lccc}
\hline
Lyr   & 8b   & 4b    & BW & Lyr   & 8b   & 4b    & BW\\
\hline
c1    &  0.02 & -0.05 & 32 & 2.1c2 &  0.00 &  0.14 &  4\\
1.0c1 & -0.01 &  0.04 &  4 & 2.2c1 & -0.01 & -0.07 &  4\\
1.0c2 & -0.01 & -0.08 &  4 & 2.2c2 & -0.01 & -0.04 &  4\\
1.1c1 &  0.01 & -0.04 &  4 & 3.0c1 &  0.00 &  0.04 & 32\\
1.1c2 &  0.00 & -0.01 &  4 & 3.0c2 & -0.03 & -0.03 & 32\\
1.2c1 & -0.01 &  0.02 &  4 & 3.1c1 &  0.01 &  0.03 & 32\\
1.2c2 &  0.00 &  0.01 &  4 & 3.1c2 &  0.00 &  0.03 & 32\\
2.0c1 & -0.02 & -0.01 &  4 & 3.2c1 &  0.02 & -0.06 & 32\\
2.0c2 &  0.00 & -0.14 &  4 & 3.2c2 & -0.01 &  0.00 & 32\\
2.1c1 & -0.01 &  0.04 &  4 & lin   &  0.00 &  0.06 & 32\\
\hline
\end{tabular}
\end{minipage}
\end{table}

\section{Related Work}

\subsection{Classical Quantization Approaches}

Early quantization work established post-training quantization (PTQ) and quantization-aware training (QAT) as the primary strategies for reducing model precision while maintaining accuracy \cite{b2}. DoReFa-Net demonstrated low-bitwidth training by quantizing weights, activations, and gradients \cite{b5}. These methods typically apply uniform quantization across all layers, which simplifies deployment but ignores heterogeneous layer characteristics.

\subsection{Mixed-Precision and Hardware-Aware Methods}

Mixed-precision quantization assigns different bit-widths to different layers. HAQ uses reinforcement learning to search for per-layer bit-widths that minimize latency on target hardware \cite{b3}, while HAWQ leverages Hessian information to determine layer sensitivity and assign precision accordingly \cite{b4}. More recent work like QuantNAS combines neural architecture search with quantization-aware training to jointly optimize architecture and quantization policy \cite{b6}. SimQ-NAS simultaneously searches for both network architecture and quantization policy \cite{b7}, and GMPQ-TE uses topological entropy to measure layer sensitivity for efficient mixed-precision allocation \cite{b8}.

These methods optimize for hardware-specific metrics such as latency, energy consumption, or model size. While effective, they do not directly target arithmetic intensity as an optimization objective. HAQ and HAWQ measure latency on target devices but do not use the Roofline model's AI metric to guide quantization decisions.

\subsection{Large Language Model Quantization}

Recent work on LLM quantization has developed specialized techniques for transformer architectures. GPTQ applies layer-wise quantization using inverse Hessian information for weight-only quantization \cite{b9}, while AWQ preserves activation magnitudes by protecting salient weight channels \cite{b10}. SmoothQuant addresses activation outliers by migrating quantization difficulty from activations to weights \cite{b11}, and INT-FlashAttention extends FlashAttention to INT8 for efficient attention computation \cite{b12}. FP8 quantization has emerged as a practical alternative, offering better hardware support on recent accelerators while maintaining accuracy \cite{b13}.

ATOM characterizes different quantization methods by their operational intensity and uses the Roofline model as an analytical tool \cite{b14}. However, ATOM optimizes for accuracy and model size rather than using AI as a primary objective. Similarly, comprehensive surveys on LLM inference and compression analyze roofline characteristics of quantized models \cite{b15}, \cite{b16}, \cite{b17}, but do not formulate quantization as an AI-maximization problem.

\subsection{Roofline Model and Performance Analysis}

The Roofline model provides a framework for understanding when operations are memory-bound versus compute-bound based on arithmetic intensity \cite{b1}. Recent work applies this model to analyze LLM inference bottlenecks, showing that token generation is typically memory-bound due to low batch sizes \cite{b15}. For memory-bound kernels, reducing memory traffic directly improves throughput, which motivates our approach of maximizing AI through quantization.

\subsection{Positioning of This Work}

To our knowledge, no prior work explicitly uses arithmetic intensity as the primary optimization objective for per-layer quantization policy. HAQ and HAWQ optimize latency and energy but do not model AI directly. ATOM analyzes quantization through the Roofline lens but optimizes for other objectives. Our contribution is to formulate quantization as an AI-maximization problem subject to accuracy constraints, implementing search algorithms that directly optimize the Roofline metric while empirically validating that AI gains translate to throughput improvements on memory-bound deployments.

\section{Conclusion}

In this paper, we introduced Arithmetic-Intensity-Aware Quantization (AIQ), a mixed-precision framework that explicitly optimizes a joint objective over model-level arithmetic intensity and accuracy, linking the Roofline model with the quantization process. On ResNet-20, AIQ finds mixed-precision configurations that lie on or near the accuracy–AI Pareto frontier and outperform uniform INT8/INT4 baselines in combined efficiency. On MobileNetV2, the AI gains achieved by AIQ translate into a roughly 66\% throughput increase in a memory-bound deployment while keeping accuracy within about 1 percentage point of FP32. The key lessons are that per-layer effects on AI are predictable from FLOPs and memory traffic while accuracy changes are not, wide layers tend to be quantized more aggressively because they dominate memory traffic, and simple AI-aware search strategies are enough to find strong configurations. We have largely achieved our project goal of demonstrating that arithmetic-intensity-aware mixed-precision quantization is both feasible and beneficial as we were able to define a clear objective that combines AI and accuracy, implement two concrete search algorithms over per-layer bit-widths, and show consistent empirical improvements over standard baselines on multiple architectures and hardware platforms. Next steps include extending AIQ to larger architectures, studying per-channel effects of quantization, implementing more complex search algorithms that follow the heuristics we discovered in this paper, and integrating AIQ into quantization-aware training to allow for the model to achieve better accuracy-efficiency tradeoffs.

\section*{Acknowledgments}
We thank Professor H.T. Kung for his guidance and valuable feedback throughout this project.

\end{document}